# IT-OSE: Exploring Optimal Sample Size for Industrial Data Augmentation

Mingchun Sun, Rongqiang Zhao*, Zhennan Huang, Songyu Ding, Jie Liu *Fellow, IEEE*

*Abstract*—In industrial scenarios, data augmentation is an effective approach to improve model performance. However, its benefits are not unidirectionally beneficial. There is no theoretical research or established estimation for the optimal sample size (OSS) in augmentation, nor is there an established metric to evaluate the accuracy of OSS or its deviation from the ground truth. To address these issues, we propose an information-theoretic optimal sample size estimation (IT-OSE) to provide reliable OSS estimation for industrial data augmentation. An interval coverage and deviation (ICD) score is proposed to evaluate the estimated OSS intuitively. The relationship between OSS and dominant factors is theoretically analyzed and formulated, thereby enhancing the interpretability. Experiments show that, compared to empirical estimation, the IT-OSE increases accuracy in classification tasks across baseline models by an average of 4.38%, and reduces MAPE in regression tasks across baseline models by an average of 18.80%. The improvements in downstream model performance are more stable. $ICD_{dev}$ in the ICD score is also reduced by an average of 49.30%. The determinism of OSS is enhanced. Compared to exhaustive search, the IT-OSE achieves the same OSS while reducing computational and data costs by an average of 83.97% and 93.46%. Furthermore, practicality experiments demonstrate that the IT-OSE exhibits generality across representative sensor-based industrial scenarios.

## I. Introduction

DATA augmentation is an effective approach to mitigate the scarcity of acquired signals and improve the performance of supervised learning systems [1]. It has been widely applied in various industrial scenarios for sensor signals, such as sensor monitoring and measurement, equipment maintenance and fault detection, and process quality control [2], [3]. Data augmentation in industrial scenarios can be categorized into extended augmentation and non-extended augmentation. The categorization depends on whether the information content changes to a certain degree [4]. Extended augmentation expands the support domain of the input distribution by generating new observations, thereby increasing the representational diversity and potential information content of the source. Typical extended augmentation approaches include simulation models and generative models [5]. Virtual data generated from simulation models relies on accurate physical modeling and precise parameter configuration, but is limited by model fidelity and the difficulty of representing complex systems. In real-world industrial scenarios, the systems often exhibit strong nonlinearity, high noise, and multi-source coupling. It is difficult to explicitly describe system behavior with equations, leading to deviations between virtual and real-world data distributions [6], [7]. In contrast, generative models do not rely on explicit physical laws but instead learn the underlying distribution directly from real data [8]. Generative models exhibit better flexibility and generalization capability in industrial data augmentation [9]. Mainstream generative models include generative adversarial networks (GANs), variational autoencoders (VAEs), and diffusion models [10]. GANs generate high-fidelity and detail-rich data through adversarial training but suffer from instability and mode collapse. VAEs exhibit strong interpretability and continuous latent structures, but the generated data may lack sharpness and realism. Diffusion models that are based on a probabilistic process of progressively adding and removing noise produce high-quality data with stable training dynamics [11]. Diffusion models have gradually become the mainstream generative models in industrial scenarios. To ensure the generated data aligns with specific application needs, prior knowledge is incorporated as conditions in diffusion models. The prior knowledge includes scene representations and sequential patterns [12]. Additionally, non-extended augmentation reconstructs or perturbs existing observations through linear or nonlinear transformations [13], [14]. The support of the input distribution remains unchanged, and no new independent information sources are introduced. Representative non-extended augmentation approaches include signal transformation-based strategies and interpolation-based strategies [15], [16]. These approaches improve data utilization efficiency and model generalization by enhancing robustness to noise, temporal distortions, and sensor uncertainties. Although data augmentation improves generalization and robustness in downstream tasks, the choice of sample size in data augmentation is not unidirectionally beneficial [17]. An insufficient sample size limits the model's ability to capture task-relevant variations in the input space, thereby constraining its discriminative and generalization capacity. Conversely, excessive augmentation introduces redundant samples, leading to distributional shifts, amplified label noise, or overfitting to the augmented data. The model performance in downstream tasks may degrade. Therefore, exploring the optimal sample size (OSS) for industrial data augmentation is necessary.

Existing approaches for the OSS estimation primarily include exhaustive search and empirical estimation. Exhaustive search involves manually setting multiple ratios of augmented

*Corresponding author: Rongqiang Zhao@zhaorq@hit.edu.cn.

Mingchun Sun, Rongqiang Zhao, Zhennan Huang, Songyu Ding, and Jie Liu are with the Faculty of Computing, Harbin Institute of Technology, Harbin 150001, China.

Mingchun Sun, Rongqiang Zhao, Zhennan Huang, Songyu Ding, and Jie Liu are with the National Key Laboratory of Smart Farm Technologies and Systems, Harbin 150001, China.

data to real data [18]. Multiple rounds of experiments are designed to observe the trend of model performance as a function of the sample size [19], [20]. OSS is determined at the point where model performance reaches its peak. Empirical estimation refers to directly determining the sample size of data augmentation based on domain knowledge. In other fields, empirical estimation also remains the main approach. For instance, in NVIDIA's Cosmos-Drive-Dreams [21], the ratio of real to augmented data is set to 2:1 during training to ensure that the generated data improves model performance. Other models used in autonomous driving adopt similar or approximately equal ratios. Likewise, Microsoft sets the ratio of real to generated data to 3:2 when training large-scale language models [22]. Overall, exhaustive search allows a relatively objective determination of OSS based on the datasets and models, providing quantifiable performance curves [23]. Empirical estimation leverages engineering experience in the same domain to quickly obtain OSS for a given scenario. Compared with exhaustive search, empirical estimation does not require extensive computational and data resources.

However, exhaustive search and empirical estimation have certain limitations. Firstly, the conditions or factors that determine OSS could not be clearly indicated in various industrial scenarios, which leads to poor interpretability. Secondly, there is currently no reliable theoretical approach for the OSS estimation. The potential increase in source diversity and latent information during data augmentation is generally not considered. The model performance in downstream tasks cannot be improved steadily. Moreover, there is a lack of intuitive evaluation metrics. The accuracy of the OSS estimation and its deviation from the ground truth cannot be intuitively evaluated. Thirdly, the generality across sensor-based industrial scenarios of the approaches in data augmentation is limited. Exhaustive search requires multiple rounds of training and testing experiments for different generated sample sizes. This incurs high computational and data costs, especially for large-scale industrial data or complex models. Empirical estimation relies on domain knowledge, the strong subjectivity results in low determinacy for the estimation. Therefore, it is essential to address these limitations.

The motivation is to explore a theoretically derived approach for reliable OSS estimation, which identifies dominant factors in the estimation, and to provide an intuitive evaluation for the estimated OSS. Exhaustive search experiments and domain-specific experience are not required. To address these issues, we propose an information-theoretic optimal sample size estimation (IT-OSE) and an interval coverage and deviation (ICD) score for exploring OSS in industrial data augmentation. The main contributions are summarized as follows:

1) We theoretically analyze and formulate the relationship between OSS and dominant factors, including the given dataset, the baseline model, and the generator. The interpretability of OSS is enhanced.
2) We propose an IT-OSE to provide reliable OSS in industrial data augmentation without relying on exhaustive search experiments and domain-specific experience. The improvements in downstream model performance are more stable. And a two-component metric ICD score is proposed to evaluate the estimated OSS intuitively.
3) We conduct extensive experiments in industrial scenarios, including the detection of rice quality in food industry and the measurement of potassium ion concentration in chemical industry. Experimental results demonstrate that, compared to empirical estimation, the IT-OSE enhances the determinacy. Compared to exhaustive search, the IT-OSE achieves the same OSS while reducing computational and data costs. Practicality experiments further demonstrate that the IT-OSE exhibits generality across representative sensor-based industrial scenarios.

## II. METHODOLOGY

In industrial data augmentation, extended augmentation is mainly used for classification tasks, as sufficient samples in each class could support the training of generative models. Non-extended augmentation is mainly used for regression tasks, where each label typically has only a limited number of samples. The IT-OSE covers the classification and regression tasks in industrial scenarios. In the IT-OSE, the information-theoretic limit-based estimation (ITLE) is proposed for extended augmentation and the minimal generalization error-based estimation (MGEE) is proposed for non-extended augmentation. The ICD score is proposed for intuitive evaluation.

### A. ITLE for Extended Data Augmentation

Extended augmentation expands the support domain of the input distribution by generating new observations. The diversity of feature representations is enhanced, leading to an increase in the potential information. The model's insufficient representation for the dataset distribution results in an information gap. Therefore, the ITLE is proposed to estimate OSS required to fill the information gap between the given dataset and the baseline model.

The OSS estimation is established on three given factors: the given dataset $\mathcal{D}_c$, the baseline model $f_\theta(\cdot)$, and the generator $\mathcal{G}_c$. Firstly, empirical mutual information lower bound $I_{lb}$ of $\mathcal{D}_c$ is estimated. To ensure the reliability of testing and to prevent information leakage, $\mathcal{D}_c$ is divided into a training set $(\boldsymbol{X}_c^{tr}, \boldsymbol{y}_c^{tr})$ and a test set $(\boldsymbol{X}_c^{test}, \boldsymbol{y}_c^{test})$. $N_c^{size}$ is the size of $\mathcal{D}_c$. Data augmentation is performed on $(\boldsymbol{X}_c^{tr}, \boldsymbol{y}_c^{tr})$ and does not involve $(\boldsymbol{X}_c^{test}, \boldsymbol{y}_c^{test})$ to prevent information leakage. The ITLE is also estimated on $(\boldsymbol{X}_c^{tr}, \boldsymbol{y}_c^{tr})$. $I_{lb}$ serves as a computable surrogate of the information-theoretic limit between the input $\boldsymbol{X}$ and the label $\boldsymbol{Y}$. Sensor signals in industrial scenarios are typically high-dimensional, nonlinear, and characterized by unknown distributions, with complex inter-dimensional mutual dependencies. To avoid the intractability of directly estimating high-dimensional probability densities, Kraskov–Stögbauer–Grassberger (KSG) nearest-neighbor estimator is used. The samples in $(\boldsymbol{X}_c^{tr}, \boldsymbol{y}_c^{tr})$ are denoted as $\{(\boldsymbol{x}_i, \boldsymbol{y}_i)\}_{i=1}^{N_c^{tr}}$. $N_c^{tr}$ is the sample size of $(\boldsymbol{X}_c^{tr}, \boldsymbol{y}_c^{tr})$. Given a neighborhood size $k$, the KSG estimated on the bootstrap subset $S$ is shown in Eq. (1).

$$\widehat{I}_{KSG}^{(b)} = \psi(k) + \psi(m) - \frac{1}{m} \sum_{i \in S^{(b)}} \left[ \psi(n_x(i) + 1) + \psi(n_y(i) + 1) \right] \quad (1)$$

$\psi(\cdot)$ denotes the digamma function, $n_x(i)$ and $n_y(i)$ represent the marginal neighbor counts within the joint k-nearest-neighbor radius centered at sample $i$. $m$ represents the sample size of $S$. $S^{(b)}$ denotes the b-th bootstrap subset sampled from $(\bm{X}_c^{\text{tr}}, \bm{y}_c^{\text{tr}})$. $k$ controls the local scale of KSG, thereby affecting the central position of the OSS interval. A collection of estimates $\{\widehat{I}_{\text{KSG}}^{(1)}, \ldots, \widehat{I}_{\text{KSG}}^{(B)}\}$ is obtained by performing $B$ bootstrap resamplings on $(\bm{X}_c^{\text{tr}}, \bm{y}_c^{\text{tr}})$. $B$ determines the conservativeness of $I_{\text{lb}}$, thereby influencing the confidence width of the OSS interval. $I_{\text{lb}}$ computes as an approximation of mutual information $I(\bm{X}; \bm{Y})$, and is defined as a conservative percentile $\mathcal{P}$. Due to the reduction in the effective information dimension of high-dimensional features constrained by physical limitations, $I_{\text{lb}}$ remains stable even in high-dimensional mutual information estimation, as shown in Eq. (2).

$$I_{\text{lb}} = \mathcal{P}\big(\{\widehat{I}_{\text{KSG}}^{(b)}\}_{b=1}^{B}, p\big), \quad \text{where } p = 5\% \tag{2}$$

Secondly, $f_\theta(\cdot)$ is trained on $(\bm{X}_c^{\text{tr}}, \bm{y}_c^{\text{tr}})$. A validation set $(\bm{X}_c^{\text{val}}, \bm{y}_c^{\text{val}})$ is partitioned from $(\bm{X}_c^{\text{tr}}, \bm{y}_c^{\text{tr}})$. Early stopping and grid search are used to obtain the optimal parameters $\theta_0$ and avoid overfitting. The objective function is shown in Eq. (3).

$$\bm{\theta}_0 = \arg\min_{\bm{\theta}} \mathcal{L}_c\big(f_{\bm{\theta}}(\bm{X}_c^{\text{val}}), \bm{y}_c^{\text{val}}\big) \tag{3}$$

$\mathcal{L}_c$ denotes the task-specific loss function. For classification tasks, the loss function is typically the cross-entropy loss. After training, $\bm{\theta}_0$ is stored as $f_{\bm{\theta}_0}$.

Thirdly, the necessity of data augmentation is determined by comparing the test error $\hat{e}_{\text{test}}$ with the information-theoretic error lower reference $p_e^{\text{lb}}$. $f_{\bm{\theta}_0}$ is evaluated on $(\bm{X}_c^{\text{test}}, \bm{y}_c^{\text{test}})$ to obtain $\hat{e}_{\text{test}}$. $\hat{e}_{\text{test}}$ quantifies the performance of $f_{\bm{\theta}_0}$, as shown in Eq. (4).

$$\hat{e}_{\text{test}} = 1 - \text{Accuracy}\big(f_{\bm{\theta}_0}(\bm{X}_c^{\text{test}}), \bm{y}_c^{\text{test}}\big) \tag{4}$$

Fano's inequality maps $I_{\text{lb}}$ and the label entropy $H(\bm{Y})$ to $p_e^{\text{lb}}$, as shown in Eq. (5).

$$p_e^{\text{lb}} = \frac{H(\bm{Y}) - I_{\text{lb}} - 1}{\log(|\mathcal{Y}| - 1)} \tag{5}$$

$|\mathcal{Y}|$ represents the number of classes, and $\log(|\mathcal{Y}| - 1)$ corresponds to the number of possible erroneous classes. The constant term 1 arises from the small-error approximation in Fano's inequality. The comparison between $\hat{e}_{\text{test}}$ and $p_e^{\text{lb}}$ is used to define and quantify the model's information gap, as shown in Eq. (6).

$$Bias = \max\{0, \hat{e}_{\text{test}} - p_e^{\text{lb}}\} \tag{6}$$

$Bias$ represents the comparison. It also serves as the generalization gap of $f_{\bm{\theta}_0}$ relative to the theoretical optimum on $\mathcal{D}_c$ and the target for augmentation. $Bias$ close to 0 indicates that $f_{\bm{\theta}_0}$ has fully utilized the information in $(\bm{X}_c^{\text{tr}}, \bm{y}_c^{\text{tr}})$, suggesting that additional augmented data are unnecessary. On the contrary, $Bias$ greater than 0 indicates the presence of an information gap that needs to be filled by augmented data.

Fourthly, Rademacher complexity $\hat{R}$ estimates the augmented sample size to fill the information gap. For classification tasks, the decision boundary is typically implemented through a sign function or a softmax mapping. $\hat{R}$ measures the expressive capacity of a model family from the perspective of empirical risk minimization. The expressive capacity directly determines the scale of the function space that the model family can achieve. Therefore, $\hat{R}$ is used to quantify the model family complexity, as shown in Eq. (7).

$$\hat{R} = \mathbb{E}_\sigma \left[ \sup_{f \in \mathcal{F}} \frac{1}{N_c^{\text{tr}}} \sum_{i=1}^{N_c^{\text{tr}}} \sigma_i f(\bm{x}_i) \right] \tag{7}$$

$\mathcal{F}$ denotes the hypothesis space spanned by $f_{\bm{\theta}}(\cdot)$. $\sigma_i \in \{\pm 1\}^{N_c^{\text{tr}}}$ represents the $j$-th independently generated Rademacher random variable, and $f(\bm{x}_i)$ denotes a real-valued model response in $\mathcal{F}$ on the $i$-th training sample. $\mathbb{E}_\sigma[\cdot]$ denotes the expectation taken over all possible values of $\sigma_i$, which makes the complexity estimation more stable. For each $\bm{\sigma}^{(j)}$, the maximum model response is computed as a weighted average, which is averaged over $M$ iterations to obtain an estimate of the empirical complexity on $(\bm{X}_c^{\text{tr}}, \bm{y}_c^{\text{tr}})$. $\hat{R}$ is scaled according to the sample size to facilitate comparisons across different sample scales. Empirical complexity scaling factor $\kappa_{\text{emp}}$ is shown in Eq. (8).

$$\kappa_{\text{emp}} = \hat{R} \cdot \sqrt{N_c^{\text{tr}}} \tag{8}$$

At a confidence level of $1 - \delta$, the inverse function of Rademacher generalization bound calculates the theoretical sample size $n_{\text{eff}}$ for $\mathcal{D}_c$ and $f_{\bm{\theta}}(\cdot)$, as shown in Eq. (9).

$$n_{\text{eff}} = \frac{\big(C \cdot \kappa_{\text{emp}} + \sqrt{\frac{\log(1/\delta)}{2}}\big)^2}{\gamma^2} \tag{9}$$

$\gamma$ denotes the target tolerance, which is used to determine the saturation threshold for generalization error. Due to the highly non-independent and redundant parameterization in models, a portion of degrees of freedom that do not contribute to computation cannot be effectively eliminated. The bound on $\hat{R}$ is relatively loose, and $C$ compensates for the looseness by encapsulating the Lipschitz constant of the loss function and the scaling factor. OSS is then calculated in Eq. (10).

$$N_c = \max(0, n_{\text{eff}} - N_c^{\text{tr}}) \tag{10}$$

It is worth noting that even when $Bias$ is not 0, $N_c$ may also be 0. A zero $N_c$ indicates that the model has saturated on the current dataset, suggesting that data augmentation is unnecessary. It also indicates a model saturation regime, where the baseline model capacity is sufficient relative to the scale and information content of the given dataset. The dual constraints of Eq. (6) and Eq. (10) ensure the accuracy of estimations. The theoretical effective sample size required to achieve the target generalization tolerance is already satisfied by the real training data. The computation of $N_c$ is conservative and deviates from the ground truth. A correction factor $\alpha$ is proposed to account for deviations between theoretical estimates and the ground truth. It is calculated as the product of four interpretable components, each driven by computable quantities rather than tuned heuristically, as shown in Eq. (11).

$$\alpha = \alpha_{\text{R}} \cdot \alpha_{\text{bound}} \cdot \alpha_{\text{opt}} \cdot \alpha_{\text{loss}} \tag{11}$$

$\alpha_{\text{R}}$ represents a confidence adjustment for the empirical Rademacher distribution based on the variance across repetitions, using a high-percentile (95th) ratio to account for uncertainty. $\alpha_{\text{bound}}$ accounts for relaxation in the theoretical bound under multi-class settings. It is set as a mild logarithmic function of the number of classes. $\alpha_{\text{opt}}$ adjusts for the ratio between spectral complexity and empirical complexity of the model. It is determined by the ratio between spectral complexity and empirical complexity. $\alpha_{\text{loss}}$ corrects for surrogate error arising from different loss functions. It is fixed according to the loss function used. Together, the factors provide a conservative upper bound for the OSS estimation. $\alpha$ is used to adjust $N_c$, yielding the upper bound of OSS $N_c^{\text{upper}}$. OSS is reported as an interval $[N_c, N_c^{\text{upper}}]$. The interval is clipped to a narrow range to ensure conservative yet stable OSS estimation, where $N_c^{\text{upper}}$ is shown in Eq. (12).

$$N_c^{\text{upper}} = \alpha \cdot N_c \tag{12}$$

Finally, information contribution ratio $\rho$ is quantified to estimate OSS. $\rho$ between the real data $\boldsymbol{X}_r$ and the augmented data $\boldsymbol{X}_g$ represents the number of $\boldsymbol{X}_g$ that are information-theoretically equivalent to $\boldsymbol{X}_r$. $\rho$ is independent of $\mathcal{D}_c$ and $f_{\theta_0}$, it depends solely on $\mathcal{G}_c$. To quantify $\rho$, $\boldsymbol{X}_r$ and $\boldsymbol{X}_g$ are reduced in dimensionality using principal component analysis (PCA) and Min–Max normalization to improve the stability of mutual information estimation. Next, the KSG estimates the mutual information $I(\boldsymbol{X}_r; \boldsymbol{X}_g|\boldsymbol{Y})$ between $\boldsymbol{X}_r$ and $\boldsymbol{X}_g$ strictly under the premise of consistent category labels. The joint input entropy $H(\boldsymbol{X}_r|\boldsymbol{Y})$ of $\boldsymbol{X}_r$ is computed in a discrete manner. $\rho$ is computed as shown in Eq. (13).

$$\rho = 1 - \xi \cdot \frac{I(\boldsymbol{X}_r; \boldsymbol{X}_g|\boldsymbol{Y})}{H(\boldsymbol{X}_r|\boldsymbol{Y}) + \varepsilon}, \quad \text{where } \xi = \exp\left(-\frac{\mathcal{K}}{\tau}\right) \tag{13}$$

$\varepsilon$ is a numerical stability term. $\xi$ is a generator-dependent complexity factor. It precisely distinguishes the distributional expansion brought about by high-complexity models from the redundant overlaps caused by low-strategy transformations. $\mathcal{K}$ is the parameter count of $\mathcal{G}_c$ and $\tau$ is a scaling factor. $[N_c, N_c^{\text{upper}}]$ is then transformed into the estimated interval $[\overline{N_c}, \overline{N_c^{\text{upper}}}]$. A reliable interval estimation rather than an overconfident point estimate provided, as shown in Eq. (14).

$$\overline{N_c}, \overline{N_c^{\text{upper}}} = \frac{N_c}{\rho}, \frac{N_c^{\text{upper}}}{\rho} \tag{14}$$

The ITLE identifies the information gap by comparing $p_e^{\text{lb}}$ with the model's performance, and derives OSS for model performance improvements. The width of the OSS interval reflects the inherent uncertainty in real-world scenarios, especially in small-sample industrial scenarios where uncertainty is stronger. Furthermore, the ITLE analyzes and formulates that three dominant factors influence the estimation: the given dataset, the baseline model, and the generator. The algorithm flow of the ITLE is described in Algorithm 1.

---
**Algorithm 1** ITLE
**Require:** $\mathcal{D}_c, f_\theta(\cdot), \mathcal{G}_c$
**Ensure:** OSS
1: $I_{\text{lb}} \leftarrow \mathcal{D}_c$ using Eqs. (1) - (2)
2: $f_\theta(\cdot) \leftarrow f_{\theta_0}$ using Eq. (3)
3: $Bias \leftarrow \mathcal{D}_c, f_\theta(\cdot)$ using Eqs. (4) - (6)
4: **if** $Bias > 0 + \epsilon, \epsilon \to 0$ **then**
5: $\quad [N_c, N_c^{\text{upper}}] \leftarrow \mathcal{D}_c, f_\theta(\cdot), \mathcal{G}_c$ using Eqs. (7) - (12)
6: $\quad [\overline{N_c}, \overline{N_c^{\text{upper}}}] \leftarrow \mathcal{G}_c$ using Eqs. (13) - (14)
7: **else**
8: $\quad [\overline{N_c}, \overline{N_c^{\text{upper}}}] \leftarrow [0, 0]$
9: **end if**
10: OSS $\leftarrow [\overline{N_c}, \overline{N_c^{\text{upper}}}]$
11: **return** OSS

---

### B. MGEE for non-extended data augmentation

Non-extended data augmentation reconstructs or perturbs existing samples via linear or nonlinear transformations, without extending the informational support of the input distribution or introducing new independent information sources. Given that the information content remains essentially unchanged, model performance is primarily constrained by generalization error. Therefore, the MGEE is proposed to estimate OSS to mitigate the baseline model's generalization error.

Similarly, the estimation in the MGEE is established on three given factors: the given dataset $\mathcal{D}_r$, the baseline model $f_\eta(\cdot)$, and the generator $\mathcal{G}_r$. Firstly, $f_\eta(\cdot)$ is trained on $\mathcal{D}_r$, and the effective model complexity $\mathcal{A}_{\text{PB}}$ is estimated. $N_r^{\text{size}}$ is the sample size of $\mathcal{D}_r$. Similar to the ITLE, $\mathcal{D}_r$ is split into a training set $(\boldsymbol{X}_r^{\text{tr}}, \boldsymbol{y}_r^{\text{tr}})$ and a test set $(\boldsymbol{X}_r^{\text{test}}, \boldsymbol{y}_r^{\text{test}})$ to prevent information leakage. The MGEE is estimated using $(\boldsymbol{X}_r^{\text{tr}}, \boldsymbol{y}_r^{\text{tr}})$. A validation set $(\boldsymbol{X}_r^{\text{val}}, \boldsymbol{y}_r^{\text{val}})$ is carved out from $(\boldsymbol{X}_r^{\text{tr}}, \boldsymbol{y}_r^{\text{tr}})$. Early stopping and model selection are used to obtain the optimal weights $\eta_0$, as shown in Eq. (15).

$$\eta_0 = \arg\min_{\eta} \mathcal{L}_r\big(f_\eta(\boldsymbol{X}_r^{\text{val}}), \boldsymbol{y}_r^{\text{val}}\big) \tag{15}$$

$\mathcal{L}_r$ denotes the task-specific loss function. After training, $\eta_0$ is saved, denoted as $f_{\eta_0}$. In regression tasks, model outputs continuous real-valued variables, and prediction errors could be assumed to approximately follow a Gaussian distribution. It is further assumed that the posterior distribution of the model parameters $Q$ could also be approximated as Gaussian. PAC-Bayes theory measures the deviation of the model from prior assumptions in parameter space by comparing the posterior $Q$ with the prior $P$ via the Kullback–Leibler (KL) divergence. The distribution of $P$ is $\mathcal{N}(0, \sigma_p^2)$, it is a non-data-dependent prior. PAC-Bayes theory is applicable not only to nonlinear neural networks but also to models with uncertainty under finite sample sizes, making it suitable to quantify the model complexity. On $\mathcal{D}_r$, $\mathcal{A}_{\text{PB}}$ is computed using PAC-Bayes theory, as shown in Eq. (16).

$$\mathcal{A}_{\text{PB}} = D_{\text{KL}}(Q\|P) \approx \sum_j \left[\log\frac{\sigma_p}{\sigma_q^{(j)}} + \frac{(\sigma_q^{(j)})^2 + (w_j)^2}{2\sigma_p^2} - \frac{1}{2}\right] \tag{16}$$

$D_{\text{KL}}(Q\|P)$ is the KL divergence between $Q$ and $P$, which is used for quantifying the difference in parameter space. $Q$ is an approximate distribution of the trained parameters, with each parameter $w_j$ represented by its point estimate and empirical standard deviation $\sigma_q^{(j)}$. $\mathcal{A}_{\text{PB}}$ captures the information-theoretic measure of complexity on $\mathcal{D}_r$ and serves as the foundation for next steps.

Secondly, theoretical generalization error $G(a)$ is calculated. $G(a)$ represents the expected average prediction error of the model on unseen data after being trained on $\mathcal{D}_r$ combined with augmented data at ratio $a$. In the MGEE, $G(a)$ serves as a quantitative measure of the model's theoretical generalization capability. Similarly, effective sample size $n_{\text{eff}}(a)$ is calculated by $1-\rho$, representing the information redundancy coefficient of $\boldsymbol{X}_g$ relative to $\boldsymbol{X}_r$. $1-\rho$ depends solely on $\mathcal{G}_r$, and its computation is shown in Eq. (13). For the continuous $a$, $n_{\text{eff}}(a)$ is calculated, as shown in Eq. (17).

$$n_{\text{eff}}(a) = N_r^{\text{tr}} \frac{1+a}{1+a\cdot(1-\rho)}, \quad \text{where } a = \frac{N_{\text{aug}}}{N_r^{\text{tr}}} \tag{17}$$

$N_r^{\text{tr}}$ is the sample size of $(\boldsymbol{X}_r^{\text{tr}}, \boldsymbol{y}_r^{\text{tr}})$. Different $\mathcal{G}_r$ vary in their ability to capture and reproduce the real data distribution, which results in differing degrees of information overlap between the augmented data and the real data. $1 - \rho$ depends on $\mathcal{G}_r$. $n_{\text{eff}}(a)$ is used to approximate $G(a)$ based on the PAC-Bayesian generalization bound, as shown in Eq. (18).

$$G(a) = \sqrt{\frac{\mathcal{A}_{\text{PB}}}{n_{\text{eff}}(a)}} \quad (18)$$

Thirdly, optimal augmentation ratio $a^*$ is determined by $G(a)$. The curve of $G(a)$ reflects how the model's generalization performance evolves and saturates as $a$ increases. When the samples are augmented but the model's loss no longer decreases, the derivative of $G(a)$ with respect to $a$ is theoretically equal to 0. By requiring the derivative of $G(a)$ with respect to $a$ approaches 0 ($\leq \iota$), $a^*$ corresponding to the point where the model performance begins to saturate could be determined. The augmented sample size corresponding to $a^*$ is also calculated, as shown in Eq. (19) and Eq. (20).

$$a^* = \arg\min_a \left| \frac{dG(a)}{da} \right| \approx \left\{ a \left| \frac{dG(a)}{da} \right| \leq \iota \right\} \quad (19)$$

$$N_r = N_r^{\text{tr}} \cdot a^* \quad (20)$$

Finally, $N_r$ is adjusted to obtain OSS. The estimation of $N_r$ is conservative. To refine the estimation, the correction factor $\beta$ is proposed to compensate for uncertainties arising from multiple sources. Similar to $\alpha$, each of the three interpretable components is driven by computable quantities rather than tuned heuristically, as shown in Eq. (21).

$$\beta = \beta_{\text{PAC}} \cdot \beta_\rho \cdot \beta_{\text{emp}} \quad (21)$$

Complexity correction factor $\beta_{\text{PAC}}$ compensates for the influence of model complexity relative to the parameter number. It is calculated using the smooth hyperbolic tangent function. Redundancy correction factor $\beta_\rho$ adjusts for the impact of redundant augmented data on the effective sample size. It is adjusted by the quadratic term $\rho^2$. Empirical stability correction factor $\beta_{\text{emp}}$ accounts for the fluctuations of empirical errors, which is computed based on the relative standard deviation from multiple augmentation experiments. The upper bound $N_r^{\text{upper}}$ is obtained by $N_r$ and $\beta$, as shown in Eq. (22).

$$N_r^{\text{upper}} = N_r \cdot \beta \quad (22)$$

Therefore, OSS interval $[N_r, N_r^{\text{upper}}]$ is estimated by the MGEE for regression tasks. The MGEE determines OSS by identifying the point at which $G(a)$ ceases to decrease with increasing $a$. The improvement in model performance gradually approaches saturation in $[N_r, N_r^{\text{upper}}]$. Moreover, the MGEE analyzes and formulates the three dominant factors, including the given dataset, the baseline model, and the generator. The algorithm flow of the MGEE is described in Algorithm 2.

---
**Algorithm 2** MGEE
**Require:** $\mathcal{D}_r, f_\eta(\cdot), \mathcal{G}_r$
**Ensure:** OSS
1: $f_\eta(\cdot) \leftarrow f_{\eta_0}$ using Eq. (15)
2: $\mathcal{A}_{\text{PB}} \leftarrow \mathcal{D}_r, f_\eta(\cdot)$ using Eq. (16)
3: $G(a) \leftarrow \mathcal{D}_c, f_\theta(\cdot), \mathcal{G}_r$ using Eqs. (17) - (18)
4: $N_r \leftarrow G(a)$ using Eqs. (19) - (20)
5: $[N_r, N_r^{\text{upper}}] \leftarrow \mathcal{D}_c, f_\theta(\cdot), \mathcal{G}_r$ using Eqs. (21) - (22)
6: OSS $\leftarrow [N_r, N_r^{\text{upper}}]$
7: **return** OSS

---

### C. ICD Score for Evaluation

The two-component evaluation metric ICD score is proposed to qualitatively and quantitatively evaluate the deviation of OSS estimated by the IT-OSE from the ground truth $\tilde{N}_{\text{true}}$. Firstly, to eliminate the influence of differences in sample size magnitudes across datasets on the evaluation, the sample size of the dataset is used to construct a reference quantile $Q(N)$. $N$ denotes the sample size. $Q(N)$ provides a unified reference point across different datasets, as shown in Eq. (23).

$$Q(N) = 10^{\phi(\log_{10} N)}, \quad \text{where } \phi = \text{round}(\cdot) \quad (23)$$

round$(\cdot)$ denotes the function that rounds to the nearest integer. The comparability between datasets of different scales is established. The metric computation is not affected by absolute differences in sample size, but rather reflects the accuracy of the estimate relative to the dataset scale.

Secondly, the qualitative evaluation $ICD_{\text{cov}}$ is conducted to determine whether OSS estimated by the IT-OSE covers $\tilde{N}_{\text{true}}$. Due to the uncertainty of the augmented data, $N_{\text{true}}$ was used to replace $\tilde{N}_{\text{true}}$ in the calculations. $N_{\text{true}}$ was calculated by $Q(N)$, as shown in Eq. (24).

$$N_{\text{true}} = \arg\min_{q \in \boldsymbol{\mathcal{Q}}} |q - \tilde{N}_{\text{true}}|, \quad \boldsymbol{\mathcal{Q}} = \{ k \cdot Q(N) \mid k \in \mathbb{Z}_{\geq 0} \} \quad (24)$$

If the OSS interval contains $N_{\text{true}}$, $ICD_{\text{cov}}$ is assigned to 1. Otherwise, it is 0, as shown in Eq. (25).

$$ICD_{\text{cov}} = \begin{cases} 1, & \text{if } N_{\text{true}} \in \left\{ [\overline{N_c}, \overline{N_c^{\text{upper}}}], [N_r, N_r^{\text{upper}}] \right\} \\ 0, & \text{otherwise} \end{cases} \quad (25)$$

Thirdly, the quantitative evaluation $ICD_{\text{dev}}$ is conducted to measure the deviation of OSS from $\tilde{N}_{\text{true}}$. Due to uncertainties inherent in data augmentation, OSS estimated by the IT-OSE provides both an upper bound and a lower bound. The midpoint of its interval is used. When the denominator $N_{\text{true}}$ is 0, $N_{\text{true}}$ is taken as an integer that is one order of magnitude smaller than $Q(N)$. $ICD_{\text{dev}}$ quantifies the deviation of OSS center relative to $N_{\text{true}}$. Taking the MGEE estimation as an example, the calculation of $ICD_{\text{dev}}$ is shown in Eq. (26).

$$ICD_{\text{dev}} = \frac{0.5 \cdot (N_r + N_r^{\text{upper}}) - N_{\text{true}}}{N_{\text{true}}} \quad (26)$$

Finally, the two-component metric ICD score integrates $ICD_{\text{cov}}$ and $ICD_{\text{dev}}$ to provide a comprehensive evaluation for the OSS estimated by the IT-OSE, as shown in Eq. (27).

$$ICD = [ICD_{\text{cov}}, ICD_{\text{dev}}] \quad (27)$$

## III. EXPERIMENTS

### A. Industrial Scenarios and Experimental Setup

*1) Industrial Scenarios and Sensors:* We conducted extensive experiments on two representative industrial scenarios. The scenarios included the detection of rice quality with olfactory sensing in food industry domain and the measurement of potassium ion concentration with electrochemical sensing in chemical industry domain. The two scenarios corresponded to classification and regression tasks, respectively. In food industry, stored rice in warehouses exhibited three quality states: normal, expired, and moldy. To prevent food loss and potential safety hazards, timely detections of expired and moldy rice were necessary. An electronic nose (e-nose, PEN3 of AIRSENSE, Germany) was used as the sensor, comprising 10 channels with different gas-sensitive materials [24], [25]. Distinct odors were captured and converted into electrical signals to characterize the rice quality states. The olfactory detection scenario was denoted as $\mathcal{IS}_1$. In chemical industry, to ensure water quality safety and environmental compliance, regular measurement of potassium ion concentration in industrial wastewater and tap water was necessary. A potassium ion-selective electrode (K-ISE, Orion9320BNWP of Thermo Fisher Scientific, USA) was used as the sensor. The K-ISE selectively sensed the activity of potassium ions in solution through its electrode membrane, generating potential changes [13]. The voltage was measured by a programmable logic controller (PLC) device and recorded by a host computer. The electrochemical measurement scenario was denoted as $\mathcal{IS}_2$. $\mathcal{IS}_1$ and $\mathcal{IS}_2$ were shown in Fig. 1.

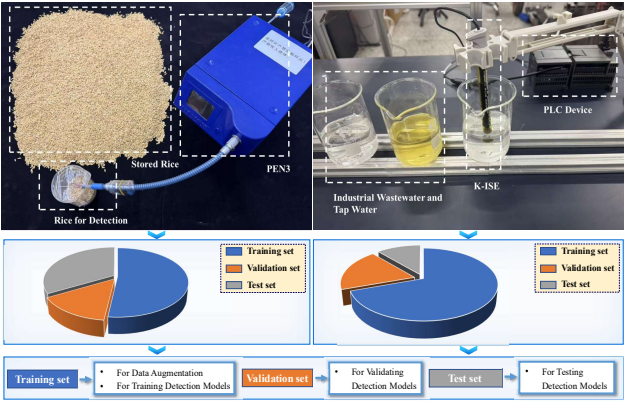

Fig. 1. Industrial scenarios and Dataset Splits: $\mathcal{IS}_1$ (left, classification task) and $\mathcal{IS}_2$ (right, regression task).

*2) Dataset Preparation:* In $\mathcal{IS}_1$, a dataset for the classification task was constructed. The dataset contained 297 samples, representing three categories: normal, expired, and moldy. To prevent information leakage, the training and test sets were acquired and constructed separately, and the sample size in each category were balanced. To ensure the generalization, the samples in the test set were acquired separately from crucial locations in the local warehouses. Each sample included 10 features representing the 10 sensor channels and 1 label representing the category. The training set contained 198 samples, and the test set contained 99 samples. The validation set was divided from the training set in a 7:2 ratio. In $\mathcal{IS}_2$, a dataset for the regression task was constructed. The dataset included 79 samples. Similarly, the training and test sets were acquired and constructed separately. Each sample included 10 features representing voltages and 1 label representing the concentration. The training set contained 70 samples, and the test set contained 9 samples. The validation set was also divided from the training set in the same ratio. All the labels were obtained from professional testing institutions. Dataset splits in $\mathcal{IS}_1$ and $\mathcal{IS}_2$ were also shown in Fig. 1.

*3) Model Structure and Parameters:* In $\mathcal{IS}_1$, commonly used models spanned a range from machine learning to deep learning, including logistic regression (LR), support vector machine (SVM), artificial neural network (ANN), and convolutional neural network (CNN) [13], [26], [27]. In addition to these models, in $\mathcal{IS}_2$, long short-term memory networks (LSTM) and Transformers were widely used for measurement due to their nonlinear feature extraction capabilities [24], [25], [28]. The models provided flexible data modeling approaches for the olfactory sensor detection and the electrochemical measurement. Therefore, the six models with varying levels of complexity were used in our experiments, including LR (logistic/linear), SVM (SVC/SVR), ANN, CNN, LSTM, and Transformer. A range of complexity from simple to deep was covered by the models. For reproducibility, the model structure and parameters used in $\mathcal{IS}_1$ and $\mathcal{IS}_2$ were shown in Table 1.

*4) Generator:* In extended augmentation, diffusion models served as $\mathcal{G}_c$ due to their superiority over GANs and VAEs in applications [8], [9]. In $\mathcal{IS}_1$, three advanced diffusion models were used, including ImageTime, Diffusion-TS, and DiffTime [10], [11], [12]. In non-extended augmentation, augmentation strategies served as $\mathcal{G}_r$. In $\mathcal{IS}_2$, four advanced strategies were used, including RIM, TSW, DTW, and Non-SIM [13], [14], [15], [16]. The main generators were ImagenTime and RIM, both of which represented state-of-the-art approaches. In Fig. 2, a real signal categorized as "Normal" was used to compare with its generated signals in $\mathcal{IS}_1$. A real signal with a concentration of 100 ppm was used for comparison with its generated signals in $\mathcal{IS}_2$. The generated data provided additional information while maintaining semantic consistency, rather than being identical to the real.

TABLE I
MODEL STRUCTURE AND PARAMETERS USED IN $\mathcal{IS}_1$ AND $\mathcal{IS}_2$

| Scenario | Model | Structure/Parameters | Activation | Optimizer/Learning Rate | Epochs/Max$_{\text{iter}}$ |
|---|---|---|---|---|---|
| $\mathcal{IS}_1$ | LR | Logistic | None | None | 200 |
| | SVC | Kernel='rbf', C=1.0 | None | None | 200 |
| | ANN | Hidden=[1024, 512, 256] | ReLU | Adam/$10^{-3}$ | 200 |
| | 1D-CNN | Kernel=3, Channels=[32,64,128] | ReLU | Adam/$10^{-3}$ | 200 |
| | LSTM | Hidden=64, FC=[128,64] | tanh/ReLU | Adam/$10^{-3}$ | 200 |
| | Transformer | d$_{\text{model}}$=128, layers=2, n$_{\text{head}}$=1, FFN=256 | ReLU | Adam/$10^{-3}$ | 200 |
| $\mathcal{IS}_2$ | LR | Linear | None | None | 100 |
| | SVR | Kernel='linear', C=1.0 | None | None | 100 |
| | ANN | Hidden=[128, 64] | ReLU | Adam/$10^{-3}$ | 100 |
| | 1D-CNN | Kernel=3, Channels=[32,64] | ReLU | Adam/$10^{-3}$ | 100 |
| | LSTM | Hidden=64, FC=[64,32] | tanh/ReLU | Adam/$10^{-3}$ | 100 |
| | Transformer | d$_{\text{model}}$=64, layers=2, n$_{\text{head}}$=1, FFN=128 | ReLU | Adam/$10^{-3}$ | 100 |

*5) Experiment Setup:* The NVIDIA GeForce RTX 3090 was used for training the diffusion models and the extended data augmentation. The 13th Gen Intel Core i7-13700F CPU

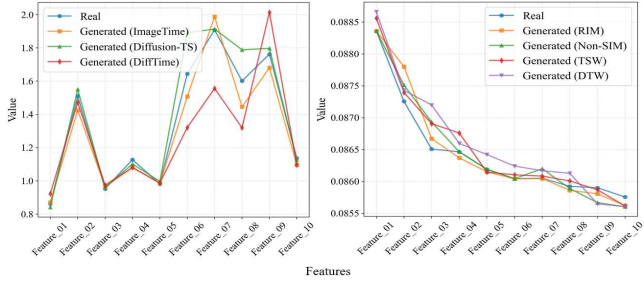

Fig. 2. Comparison between generated signals and real signal: "Normal" in $\mathcal{IS}_1$ (left) and "100ppm" in $\mathcal{IS}_2$ (right).

was used for the non-extended data augmentation and training the models in downstream tasks.

### B. Exploring OSS with IT-OSE

*1) OSS and Key Parameters in IT-OSE:* In $\mathcal{IS}_1$ and $\mathcal{IS}_2$, OSS for classification and regression tasks was estimated by the ITLE and the MGEE, respectively. The hyperparameters were determined by grid search and fixed. $K$ was set to 5, $B$ was set to 200, $\delta$ was set to 0.05, and $\gamma$ was set to 0.03 for the stability. The given dataset for each task was fixed. The baseline models were six commonly used models with different levels of complexity. Grid search was used to get optimal parameters for each model. K-fold cross-validation was used to obtain a more reliable evaluation of the model's performance. Table 2 and Table 3 showed the OSS estimations and the key parameters. $G(a)$ were the scale-transformed values of the curve evaluated at different $a$. The key parameters clarified how OSS was derived step by step. The interpretability was enhanced. The difference in complexity magnitude between the ITLE and the MGEE was attributable to the different methods used for computing complexity.

TABLE II
OSS AND KEY PARAMETERS IN ITLE

| Model | $I_{lb}$ | $p_e^{lb}$ | $\hat{e}_{\text{test}}$ | $Bias$ | $\hat{R}$ | $\alpha$ | OSS |
|---|---|---|---|---|---|---|---|
| LR | 0.832 | 0.059 | 0.091 | 0.032 | 0.001 | 1.22 | [0, 0] |
| SVC | 0.832 | 0.059 | 0.121 | 0.062 | 0.323 | 1.50 | [354, 530] |
| ANN | 0.832 | 0.059 | 0.232 | 0.163 | 0.300 | 1.22 | [273, 333] |
| 1D-CNN | 0.832 | 0.059 | 0.182 | 0.123 | 0.097 | 1.20 | [0, 0] |
| LSTM | 0.832 | 0.059 | 0.131 | 0.072 | 0.207 | 1.46 | [84, 124] |
| Transformer | 0.832 | 0.059 | 0.252 | 0.193 | 0.239 | 1.22 | [138, 168] |

TABLE III
OSS AND KEY PARAMETERS IN MGEE

| Model | $\mathcal{A}_{\text{PB}}$ | $G(a)$ | $a^*$ | $\beta$ | OSS |
|---|---|---|---|---|---|
| LR | 0.022 | 13.007 | 0 | 1.218 | [0, 0] |
| SVR | 0.001 | 14.260 | 0 | 1.224 | [0, 0] |
| ANN | 9.729 | 15.245 | 3.659 | 1.225 | [256, 314] |
| 1D-CNN | 14.977 | 12.036 | 4.211 | 1.238 | [295, 365] |
| LSTM | 21.377 | 15.288 | 4.661 | 1.226 | [326, 399] |
| Transformer | 75.521 | 12.720 | 6.767 | 1.219 | [474, 578] |

*2) Validation Experiments for IT-OSE:* To verify the reliability of the IT-OSE, the experiments in this subsection were conducted. Similar to $Q(N)$ in the ICD score, the sample size was set to be of the same order of magnitude as the percentiles of the given dataset (100). To ensure a wide validation range, percentiles from 100 to 700 (with a step of 100) were set. Cold start was used as the initialization strategy for the baseline models. The models were trained from scratch, and the performance improvements were fully attributed to the additional augmented samples. The influence of data augmentation on model performance was therefore quantified more accurately and objectively. In $\mathcal{IS}_1$, cross-entropy was used as the loss function, accuracy and F1-score were used as evaluation metrics. $Bias$ was also used to validate the estimation. In $\mathcal{IS}_2$, MSE was used as the loss function, MAPE and RMSE were used as evaluation metrics. Multiple evaluation metrics enabled a comprehensive evaluation of model performance and ensured the reliability of the experiments. In addition, repeated experiments were performed for each sample size to reduce randomness, and the average values of all metrics were recorded. The experimental results were shown in Fig. 3.

The shaded regions in Fig. 3 largely covered the ground truth. Fig. 3 demonstrated the reliability of the OSS estimation by the ITLE and the MGEE. Moreover, it was worth noting that in $\mathcal{IS}_1$, for the LR, when the sample size was excessively increased, the information introduced by the augmented samples became overly large relative to the given dataset. The insufficient model capacity led to underfitting. For the SVC, due to the implicit reliance of the "rbf" kernel on the kernel matrix and support vectors to map samples into a high-dimensional space, the complexity was greater. The results also confirmed that the choice of sample size for data augmentation was not unilaterally beneficial. And on datasets with small sample size, simpler models performed better.

*3) Evaluation of IT-OSE with ICD Scores:* To verify the enhancement of the determinacy, and to verify the intuitive evaluation of the ICD score, the experiments in this subsection were conducted. The ICD scores in the ITLE and the MGEE were shown in Table 4. Experimental results demonstrated that, through the step-by-step derivation of the ITLE and the MGEE, the OSS closely approached the ground truth. The enhancement of the determinacy was demonstrated. The coverage of the ground truth by the estimated OSS, as well as the discrepancy from the ground truth, was intuitively presented using the two-component metric $ICD_{\text{cov}}$ and $ICD_{\text{dev}}$. The intuitive evaluation was provided by the ICD Score. Moreover, the absence of increased information under non-extended augmentation led to an unstable estimation. The generalization error depended on various factors, including the model, data, and noise, with highly random noise having a significant impact on the model's performance. Although the MGEE produced lower ICD scores than the ITLE, it nevertheless yielded an accurate estimation. It was worth noting that the effective information content available for learning was limited in small sample datasets. More complex models like Transformers might have produced unstable OSS estimations.

*4) Comparisons with Existing Approaches:* To further verify the enhancement of the determinacy and the more stable improvements in downstream model performance, and to verify the reductions in computational and data costs, the experiments were conducted. In existing approaches, exhaustive search (denoted as $\mathcal{B}_{\text{es}}$) [19], [20] and empirical estimation were used as baselines. The exhaustive search was shown in Fig. 3. In empirical estimation, domain expertise from NVIDIA [21] (denoted as $\mathcal{B}_\text{N}$) and Microsoft [22] (denoted as $\mathcal{B}_\text{M}$) was used for comparisons. The ratios of real to augmented data were set to 2:1 and 3:2. Table 5 showed the main evaluation metrics [13], [28] (Accuracy and MAPE, denoted as $\mathcal{M}$), the ICD scores, the generated sample size representing the data cost (denoted as $Size$), and the computational time representing the computational cost (denoted as $T$) in $\mathcal{IS}_1$ and $\mathcal{IS}_2$.

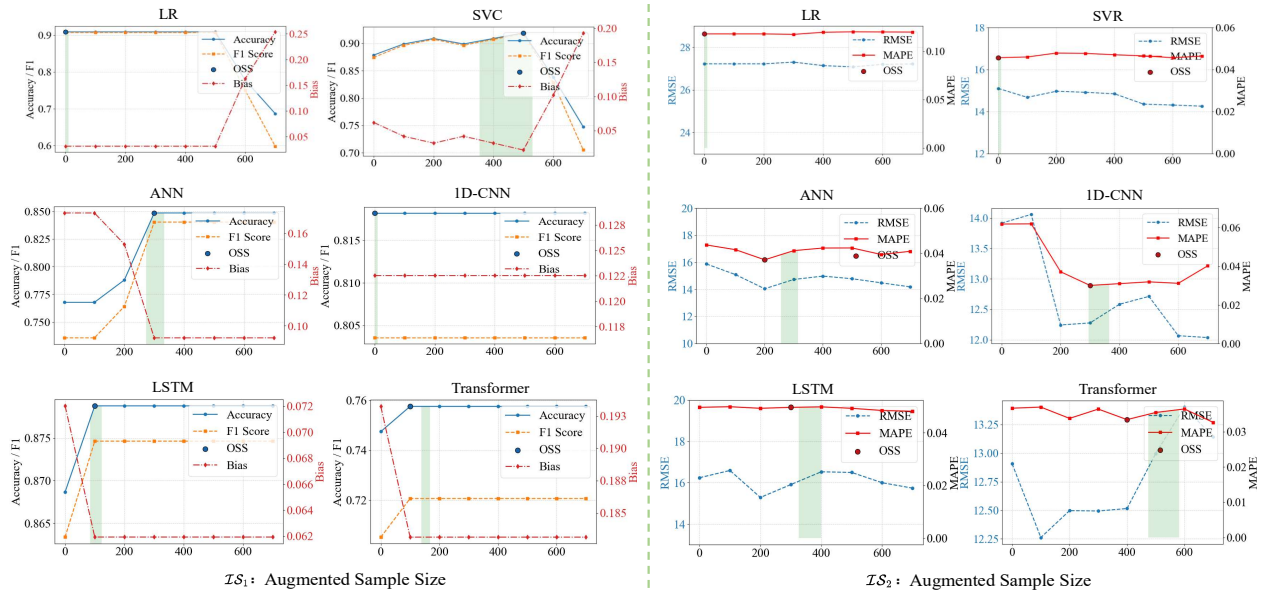

Fig. 3. Validation experiments for IT-OSE in $\mathcal{IS}_1$ and $\mathcal{IS}_2$ (shaded regions represent OSS intervals estimated by IT-OSE).

TABLE IV
ICD SCORES IN ITLE AND MGEE

| Approach | Model | OSS | $N_{\text{true}}$ | $ICD_{\text{cov}}$ | $ICD_{\text{dev}}$ | |
|---|---|---|---|---|---|---|
| ITLE | LR | [0, 0] | 0 | 1 | 0 | |
| | SVC | [354, 530] | 500 | 1 | 11.60% | |
| | ANN | [273, 333] | 300 | 1 | 1.00% | 11.60% |
| | 1D-CNN | [0, 0] | 0 | 1 | 0 | |
| | LSTM | [84, 124] | 100 | 1 | 4.00% | |
| | Transformer | [138, 168] | 100 | 0 | 53.00% | |
| MGEE | LR | [0, 0] | 0 | 1 | 0 | |
| | SVR | [0, 0] | 0 | 1 | 0 | |
| | ANN | [256, 314] | 200 | 0 | 42.50% | 17.47% |
| | 1D-CNN | [295, 365] | 300 | 1 | 10.00% | |
| | LSTM | [326, 399] | 300 | 0 | 20.83% | |
| | Transformer | [474, 578] | 400 | 0 | 31.50% | |

Since $\mathcal{B}_\text{es}$ was used to obtain $N_\text{true}$ and served as the basis for ICD score calculations, $\mathcal{B}_\text{N}$ and $\mathcal{B}_\text{M}$ were primarily used for the comparisons of $\mathcal{M}$ and ICD scores. The results showed that, compared with $\mathcal{B}_\text{N}$ and $\mathcal{B}_\text{M}$, the IT-OSE improved the average accuracy by 4.38% in $\mathcal{IS}_1$, and reduced the average MAPE by 18.80% in $\mathcal{IS}_2$. The improvements in downstream model performance were more stable. Compared to $\mathcal{B}_\text{N}$ and $\mathcal{B}_\text{M}$, the average $ICD_\text{dev}$ was reduced by 49.30% for the models in $\mathcal{IS}_1$ and $\mathcal{IS}_2$. The number of instances where $ICD_\text{cov}$ equals 1 also increased. $\mathcal{M}$ and the ICD scores further demonstrated that the enhancement of the determinacy. Moreover, the comparisons of $Size$ and $T$ indicated that, compared with $\mathcal{B}_\text{es}$, the IT-OSE achieved the same OSS while reducing $T$ by 83.97%, and reduced $Size$ by 93.46%. The computational and data costs were significantly reduced.

### C. Investigation of dominant factors's Influence

The given dataset, the baseline model, and the generator were analyzed and formulated to be dominant factors in the step-by-step derivation. This section further investigated the influences of them.

*1) Given Dataset:* To ensure that the dataset could support model training, a relatively sufficient dataset size should be retained. Given the limited scale of the datasets themselves, 90% of the datasets were randomly selected from the training set for comparisons. OSS of each model were shown in Table 6. Experimental results showed that, as the sample size increased, the usable information was increased. A larger OSS was required to further fill the remaining information gap and improve generalization performance. Therefore, the size of the given dataset is positively correlated with OSS based on the baseline models. The given datasets served as a foundation for the investigation.

*2) Baseline Model:* Complexity determined the information content that the baseline model could capture from the given dataset. Table 2 and Table 3 already showed the complexities of the baseline models. Due to the low model complexity of the machine learning models, OSS of each model was 0 on the given datasets. Deep learning models were more suitable for investigation due to their diversity and higher complexity. The relationship between the complexity of baseline models and OSS was established in the deep learning models, as shown in Fig. 4. The median of the OSS interval was used to replace itself, denoted as $Middle$. The curves in Fig. 4 demonstrated that as the complexity of the baseline model increased, the amount of information extractable from the given dataset also increased, leading to a corresponding rise in OSS. Therefore, the complexity of the baseline model is positively correlated with OSS based on the given dataset and the generator.

TABLE V
COMPARISONS OF IT-OSE AND BASELINES IN MAIN EVALUATION METRICS $\mathcal{M}$, ICD SCORES(BOLD INDICATES BEST PERFORMANCE OF $\mathcal{M}$ AND ICD SCORES), COMPUTATIONAL COST, AND DATA COST.

| Scenario | Model | IT-OSE $\mathcal{M}$ | ICD | Size | $T$(s) | $\mathcal{B}_{es}$ $\mathcal{M}$ | ICD | Size | $T$(s) | $\mathcal{B}_{N}$ $\mathcal{M}$ | ICD | Size | $T$(s) | $\mathcal{B}_{M}$ $\mathcal{M}$ | ICD | Size | $T$(s) |
|---|---|---|---|---|---|---|---|---|---|---|---|---|---|---|---|---|---|
| $\mathcal{IS}_1$ | LR | .919±.002 | **[1,0]** | 0 | <0.001 | .919±.002 | [1,0] | 2800 | <0.001 | .919±.002 | [0,12.0%] | 79 | <0.001 | .919±.002 | [0,17.0%] | 105 | <0.001 |
|  | SVC | .919±.001 | **[1,11.6%]** | 500 | <0.001 | .919±.000 | [1,0] | 2800 | <0.001 | .899±.001 | [0,80.2%] | 79 | <0.001 | .909±.001 | [0,76.7%] | 105 | <0.001 |
|  | ANN | .849±.001 | **[1,1.0%]** | 300 | 0.121 | .849±.003 | [1,0] | 2800 | 0.618 | .778±.002 | [0,70.7%] | 79 | 0.085 | .808±.001 | [0,61.0%] | 105 | 0.093 |
|  | 1D-CNN | **.818±.002** | **[1,0]** | 0 | 0.037 | .818±.002 | [1,0] | 2800 | 0.623 | .818±.002 | [0,87.34%] | 79 | 0.047 | .818±.002 | [0,90.5%] | 105 | 0.056 |
|  | LSTM | .878±.001 | **[1,4.0%]** | 100 | 0.029 | .878±.003 | [1,0] | 2800 | 0.294 | .878±.002 | [1,3.0%] | 79 | 0.029 | .878±.001 | [1,29.0%] | 105 | 0.030 |
|  | Transformer | .758±.002 | **[1,53.0%]** | 100 | 0.270 | .758±.002 | [1,0] | 2800 | 1.238 | .758±.002 | [0,12.0%] | 79 | 0.094 | .758±.002 | [1,17.0%] | 105 | 0.096 |
| $\mathcal{IS}_2$ | LR | .117±.000 | **[1,0]** | 0 | <0.001 | .117±.001 | [1,0] | 2800 | [1,0] | .116±.000 | [0,74.4%] | 35 | <0.001 | **.115±.002** | [0,71.4%] | 47 | <0.001 |
|  | SVR | .046±.001 | **[1,0]** | 0 | <0.001 | .046±.000 | [1,0] | 2800 | [1,0] | .046±.000 | [0,80.8%] | 35 | <0.001 | **.045±.001** | [0,78.7%] | 47 | <0.001 |
|  | ANN | .038±.002 | **[0,42.5%]** | 200 | 0.021 | .038±.000 | [1,0] | 2800 | [1,0] | .042±.001 | [0,80.5%] | 35 | 0.016 | .043±.001 | [0,73.5%] | 47 | 0.017 |
|  | 1D-CNN | **.030±.000** | **[1,10.0%]** | 300 | 0.043 | .030±.000 | [1,0] | 2800 | [1,0] | .063±.001 | [0,87.0%] | 35 | 0.022 | .062±.000 | [0,82.3%] | 47 | 0.022 |
|  | LSTM | .049±.000 | **[0,20.83%]** | 300 | 0.047 | .048±.000 | [1,0] | 2800 | [1,0] | .050±.000 | [0,87.0%] | 35 | 0.023 | .050±.001 | [0,82.7%] | 47 | 0.023 |
|  | Transformer | .036±.001 | **[0,31.5%]** | 400 | 0.132 | .036±.000 | [1,0] | 2800 | [1,0] | **.035±.001** | [0,90.3%] | 35 | 0.036 | .036±.002 | [0,87.0%] | 47 | 0.037 |

TABLE VI
OSS IN DIFFERENT SIZE OF GIVEN DATASET

| Scenario | Ratio | LR | SVM | ANN | LSTM | 1D-CNN | Transformer |
|---|---|---|---|---|---|---|---|
| $\mathcal{IS}_1$ | 90% | [0,0] | [158,237] | [39,47] | [0,0] | [0,0] | [128,192] |
|  | 100% | [0,0] | [354,530] | [273,333] | [84,124] | [0,0] | [138,168] |
| $\mathcal{IS}_2$ | 90% | [0,0] | [0,0] | [237,294] | [303,370] | [272,336] | [439,538] |
|  | 100% | [0,0] | [0,0] | [256,314] | [295,365] | [326,399] | [474,578] |

TABLE VIII
$\beta$ AND $Middle$ OF DIFFERENT $\rho$ IN $\mathcal{IS}_2$

| Model | Approach | OSS | $\beta$ | $Middle$ |
|---|---|---|---|---|
| ANN | RIM | [256, 314] | 1.225 | 285 |
|  | DTW | [189, 233] | 1.238 | 211 |
|  | Non-SIM | [158, 196] | 1.240 | 177 |
|  | TSW | [28, 35] | 1.251 | 32 |
| 1D-CNN | RIM | [295, 365] | 1.239 | 330 |
|  | DTW | [218, 271] | 1.246 | 245 |
|  | Non-SIM | [183, 229] | 1.250 | 206 |
|  | TSW | [39, 49] | 1.259 | 44 |
| LSTM | RIM | [326, 399] | 1.226 | 363 |
|  | DTW | [246, 304] | 1.234 | 275 |
|  | Non-SIM | [221, 273] | 1.236 | 247 |
|  | TSW | [49, 61] | 1.247 | 50 |
| Transformer | RIM | [474, 578] | 1.219 | 526 |
|  | DTW | [361, 443] | 1.235 | 402 |
|  | Non-SIM | [326, 403] | 1.237 | 365 |
|  | TSW | [91, 113] | 1.248 | 102 |

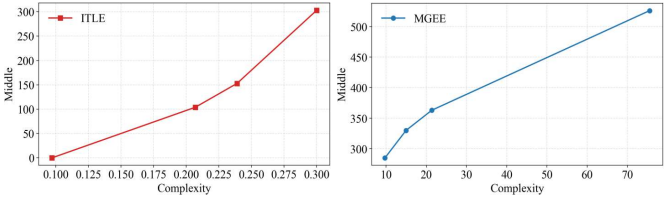

Fig. 4. Relationship between complexity of baseline models and OSS.

TABLE VII
$\rho$ AND $1-\rho$ OF AUGMENTED DATA IN $\mathcal{IS}_1$ AND $\mathcal{IS}_2$ (BOLD INDICATES BEST PERFORMANCE)

|  | $\mathcal{G}_c$ | **ImagenTime** | Diffusion-TS | DiffTime | - |
|---|---|---|---|---|---|
| $\mathcal{IS}_1$ | $\rho$ | **0.998** | 0.998 | 0.997 | - |
|  | $1-\rho$ | **0.002** | 0.002 | 0.003 | - |
|  | $\mathcal{G}_r$ | **RIM** | TSW | DTW | Non-SIM |
| $\mathcal{IS}_2$ | $\rho$ | **0.109** | 0.010 | 0.069 | 0.061 |
|  | $1-\rho$ | **0.891** | 0.990 | 0.931 | 0.939 |

*3) Generator:* $\rho$ represented $\mathcal{G}_c$ and $1-\rho$ represented $\mathcal{G}_r$, as shown in Table 7. In $\mathcal{IS}_1$, there was a high mutual dependency between $X_g$ and $X_r$, which resulted in a lower and stable $\rho$ around 1 due to informational redundancy. In contrast, in $\mathcal{IS}_2$, there were certain differences in $1-\rho$ between data augmented by transformation strategies, making $\mathcal{IS}_2$ a suitable example for investigation. Table 8 showed $\beta$ and $Middle$ of different $\mathcal{G}_r$ in $\mathcal{IS}_2$. Moreover, Fig. 5 demonstrated that as $1-\rho$ increased, $\beta$ also increased, while $Middle$ decreased. Therefore, $1-\rho$ of $\mathcal{G}_r$ and OSS were negatively correlated based on the given dataset and baseline model.

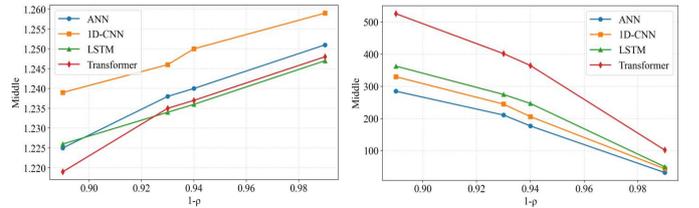

Fig. 5. Relationship between $\beta$, $Middle$ and $1-\rho$.

### D. Practicality Experiments in Industrial Scenarios

Practicality experiments were conducted to verify the generality of the IT-OSE. In $\mathcal{IS}_1$, 20 batches of rice stored at different locations in the local warehouses were detected. In $\mathcal{IS}_2$, 20 batches of wastewater and tap water from a nearby plant were measured. The best-performing baseline models from the offline experiments in each scenario were used. The ground truth was obtained from the professional testing institutions. Four approaches were compared: no data augmentation $G_1$, augmentation with the IT-OSE $G_2$, augmentation with exhaustive search $G_3$, and augmentation with an empirically determined ratio $G_4$. 3:2 was used as the empirical determined ratio. The results were shown in Table 9.

The results demonstrated that the IT-OSE exhibits generality across $\mathcal{IS}_1$ and $\mathcal{IS}_2$. Furthermore, the accurate OSS made the improvements in downstream model performances more stable. Augmentation with the IT-OSE and exhaustive search achieved same performances, but the need for extensive exploratory experiments were eliminated. Augmentation with the IT-OSE also outperformed the empirical estimation.

TABLE IX
$\mathcal{M}$ IN PRACTICALITY EXPERIMENTS (BOLD INDICATES BEST PERFORMANCE)

| Group | $\mathcal{IS}_1$ | | | | $\mathcal{IS}_2$ | | | |
|---|---|---|---|---|---|---|---|---|
| | $G_1$ | $G_2$ | $G_3$ | $G_3$ | $G_1$ | $G_2$ | $G_4$ | $G_4$ |
| $\mathcal{M}$ | 80% | **95%** | **95%** | 80% | 7.9% | **3.8%** | **3.8%** | 5.6% |

## IV. CONCLUSION

In this study, the main objective is to develop a theoretically derived approach for estimating the OSS reliably without relying on exhaustive search experiments and domain-specific expertise, and to provide an intuitive evaluation. Therefore, the IT-OSE and the ICD score are proposed. In the IT-OSE, the ITLE and the MGEE provide reliable OSS estimation for extend and non-extend augmentation, respectively. The ICD score provides an intuitive evaluation. The relationship between OSS and dominant factors is theoretically analyzed and formulated. The interpretability is enhanced. Experimental results in $\mathcal{IS}_1$ and $\mathcal{IS}_2$ demonstrate that the improvements in downstream model performance are more stable. The IT-OSE enhances the determinacy of the estimation, achieves the same OSS while reducing computational and data costs. Practicality experiments further demonstrate that the IT-OSE exhibits generality across representative sensor-based industrial scenarios.

In future research, more comprehensive dominant factors that influence the OSS estimation will be analyzed and formulated, such as noise and sampling bias. The OSS interval will be narrowed to provide a more precise estimation. Meta-learning and cross-domain transfer learning will further be incorporated as prior knowledge into the estimation, allowing augmentation approaches to be learned and transferred across similar tasks. The adaptive integration of the IT-OSE with generative models and augmentation strategies will also be explored so that the influence of generated-sample quality on the OSS estimation could be reduced. Furthermore, the IT-OSE will be extended to other industrial scenarios, such as data augmentation for harmful gas signals in industrial environments, abnormal-condition data in power systems, and defect data in industrial quality inspection. Finally, we believe that the IT-OSE is applicable in additional industrial scenarios, it could serve as the foundation of the OSS estimation in industrial data augmentation for more in-depth research.

## ACKNOWLEDGMENTS

This work was supported by the National Natural Science Foundation of China (Grant No. 62103161 and No. 62350710797), the Key Research and Development Program of Heilongjiang Province (Grant No. 2023ZX01A24 and No. JD2023GJ01-01), and the Project of Laboratory of Advanced Agricultural Sciences, Heilongjiang Province (Grant No. ZY04JD05-010).

## REFERENCES

[1] W. Ren, K. Song, C.-y. Chen, Y. Chen, J. Hong, M. Fan, X. Ouyang, Y. Zhu, and J. Xiao, "Dd-aug: A knowledge-to-image synthetic data augmentation pipeline for industrial defect detection," *IEEE Transactions on Industrial Informatics*, vol. 21, no. 3, pp. 2284–2293, 2025.
[2] L. Wang, X. Zhou, W. Ding, L. Chen, and Q. Dai, "Ensemble intrusion detection based on heterogeneous data augmentation and knowledge distillation," *IEEE Transactions on Industrial Informatics*, vol. 21, no. 11, pp. 8981–8991, 2025.
[3] G. Sun, Y. Cheng, K. Kong, Z. Zhang, and D. Zhao, "Text classification based on label data augmentation and graph neural network," *IEEE Transactions on Industrial Informatics*, vol. 21, no. 5, pp. 3966–3975, 2025.
[4] Y. Dai, Y. Xie, C. Zhang, and J. Liu, "Time-generative adversarial networks enabled ensemble prediction method for energy consumption of machine tools," *IEEE Transactions on Industrial Informatics*, vol. 21, no. 5, pp. 3796–3805, 2025.
[5] C.-X. Xu, J.-W. Xiao, and Y.-W. Wang, "A nonparametric balanced diffusion based fault diagnosis scheme for electric vehicle dc charging piles under imbalanced samples," *IEEE Transactions on Industrial Informatics*, vol. 21, no. 12, pp. 9480–9490, 2025.
[6] H. Lian, Y. Ji, M. Niu, J. Gu, J. Xie, and J. Liu, "A hybrid load prediction method of office buildings based on physical simulation database and lightgbm algorithm," *Applied Energy*, vol. 377, p. 124620, 2025.
[7] D. Q. Gbadago, S. Go, and S. Hwang, "A leap forward in chemical process design: Introducing an automated framework for integrated ai and cfd simulations," *Computers & Chemical Engineering*, vol. 192, p. 108906, 2025.
[8] X. Yang, T. Ye, X. Yuan, W. Zhu, X. Mei, and F. Zhou, "A novel data augmentation method based on denoising diffusion probabilistic model for fault diagnosis under imbalanced data," *IEEE Transactions on Industrial Informatics*, vol. 20, no. 5, pp. 7820–7831, 2024.
[9] C.-X. Xu, J.-W. Xiao, and Y.-W. Wang, "A nonparametric balanced diffusion based fault diagnosis scheme for electric vehicle dc charging piles under imbalanced samples," *IEEE Transactions on Industrial Informatics*, vol. 21, no. 12, pp. 9480–9490, 2025.
[10] I. Naiman, N. Berman, I. Pemper, I. Arbiv, G. Fadlon, and O. Azencot, "Utilizing image transforms and diffusion models for generative modeling of short and long time series," *Advances in Neural Information Processing Systems*, vol. 37, pp. 121 699–121 730, 2024.
[11] A. Coletta, S. Gopalakrishnan, D. Borrajo, and S. Vyetrenko, "On the constrained time-series generation problem," *Advances in Neural Information Processing Systems*, vol. 36, pp. 61 048–61 059, 2023.
[12] X. Yuan and Y. Qiao, "Diffusion-ts: Interpretable diffusion for general time series generation," *arXiv preprint arXiv:2403.01742*, 2024.
[13] M. Sun, R. Zhao, and J. Liu, "Enhanced potassium ion electrochemical measurement using signals augmentation and crucial features focusing," *IEEE Transactions on Instrumentation and Measurement*, vol. 74, pp. 1–15, 2025.
[14] B. K. Iwana and S. Uchida, "Time series data augmentation for neural networks by time warping with a discriminative teacher," in *2020 25th International Conference on Pattern Recognition (ICPR)*. IEEE, 2021, pp. 3558–3565.
[15] A. M. Aboussalah, M. Kwon, R. G. Patel, C. Chi, and C.-G. Lee, "Recursive time series data augmentation," in *The Eleventh International Conference on Learning Representations*, 2023.
[16] Q. Ma, Z. Zheng, J. Zheng, S. Li, W. Zhuang, and G. W. Cottrell, "Joint-label learning by dual augmentation for time series classification," in *Proceedings of the AAAI Conference on Artificial Intelligence*, vol. 35, no. 10, 2021, pp. 8847–8855.
[17] C. Bian, C. Jia, J. Li, X. Chen, and P. Wang, "Rolling bearing fault diagnosis under small sample conditions based on wdcnn-bilstm siamese network," *Scientific Reports*, vol. 15, no. 1, p. 29591, 2025.
[18] X.-P. Nguyen, S. Pandit, R. G. Reddy, A. Xu, S. Savarese, C. Xiong, and S. Joty, "Sfr-deepresearch: Towards effective reinforcement learning for autonomously reasoning single agents," *arXiv preprint arXiv:2509.06283*, 2025.
[19] S. K. Dasari, A. Cheddad, J. Palmquist, and L. Lundberg, "Clustering-based adaptive data augmentation for class-imbalance in machine learning (cada): additive manufacturing use case," *Neural Computing and Applications*, vol. 37, no. 2, pp. 597–610, 2025.

[20] Z. Gao, H. Liu, and L. Li, "Data augmentation for time-series classification: An extensive empirical study and comprehensive survey," *Journal of Artificial Intelligence Research*, vol. 83, 2025.
[21] X. Ren, Y. Lu, T. Cao, R. Gao, S. Huang, A. Sabour, T. Shen, T. Pfaff, J. Z. Wu, R. Chen *et al.*, "Cosmos-drive-dreams: Scalable synthetic driving data generation with world foundation models," *arXiv preprint arXiv:2506.09042*, 2025.
[22] M. Abdin, J. Aneja, H. Behl, S. Bubeck, R. Eldan, S. Gunasekar, M. Harrison, R. J. Hewett, M. Javaheripi, P. Kauffmann *et al.*, "Phi-4 technical report," *arXiv preprint arXiv:2412.08905*, 2024.
[23] M. V. Aragão, T. d. M. Pereira, M. d. F. Carvalho, F. A. d. Figueiredo, and S. B. Mafra, "Dynamic-balancing automl for imbalanced tabular data with adaptive resampling and complexity-aware analysis," *International Journal of Intelligent Systems*, vol. 2025, no. 1, 2025.
[24] Z. Liu, G. E. Carranza, Y. Hu, F. Wang, and M. Wong, "A gas sensor array packaged with a hierarchical neural network for gas species identification and concentration estimation," in *2023 22nd International Conference on Solid-State Sensors, Actuators and Microsystems (Transducers)*. IEEE, 2023, pp. 623–626.
[25] W. K. Tan, Z. Husin, M. L. Yasruddin, and M. A. H. Ismail, "Development of a non-destructive fruit quality assessment utilizing odour sensing, expert vision and deep learning algorithm," *Neural Computing and Applications*, vol. 36, no. 31, pp. 19 613–19 641, 2024.
[26] D. Kayali, N. A. Shama, S. Asir, and K. Dimililer, "Machine learning-based models for the qualitative classification of potassium ferrocyanide using electrochemical methods," *The Journal of Supercomputing*, vol. 79, no. 11, pp. 12 472–12 491, 2023.
[27] M. Z. Sagiroglu, E. D. Demirel, and S. Mutlu, "Accurate ion type and concentration detection using two bare electrodes by machine learning of non-faradaic electrochemical impedance measurements of an automated fluidic system," *Journal of Electroanalytical Chemistry*, vol. 961, p. 118256, 2024.
[28] I.-T. Chen, C.-C. Chen, H.-J. Dai, B. Rianto, S.-K. Huang, and C.-H. Lee, "An incremental learning method for preserving world coffee aromas by using an electronic nose and accumulated specialty coffee datasets," *IEEE Transactions on AgriFood Electronics*, vol. 2, no. 1, pp. 12–27, 2024.

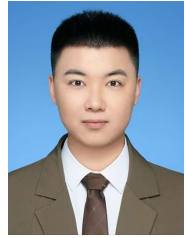
**Zhennan Huang** received the B.Eng. degree in Harbin Institute of Technology, Harbin, China, in 2025.

He is currently pursuing the M.Eng. degree with Harbin Institute of Technology, Harbin, China. His current research interests include generative models, and VLA (Vision-Language-Action) models.

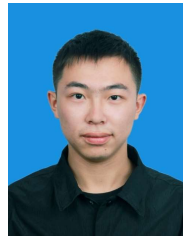
**Songyu Ding** received the B.Eng. degree from the School of Electronic Information, Wuhan University, Wuhan, China, in 2023, and the M.S. degree from the School of Electrical and Electronic Engineering, Nanyang Technological University, Singapore, in 2024.

He is currently pursuing the Ph.D. degree with Harbin Institute of Technology, Harbin, China. His current research interests include computer vision, generative models, and data augmentation.

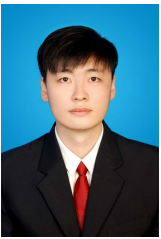
**Mingchun Sun** received the B.Eng. and M.Eng. degrees in University of Electronic Science and Technology of China, Chengdu, China, in 2018, and 2021, respectively.

He is currently pursuing the Ph.D. degree with Harbin Institute of Technology, Harbin, China. His current research interests include data augmentation and deep learning on datasets with limited samples.

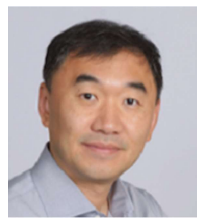
**Jie Liu** (Fellow, IEEE) received the Ph.D. degree in electrical engineering and computer science from the University of California at Berkeley, Berkeley, CA, USA, in 2001.

He is currently a Chair Professor with the Harbin Institute of Technology, Shenzhen, China. His research interests include artificial intelligence, control engineering, Internet of Things, and computer system.

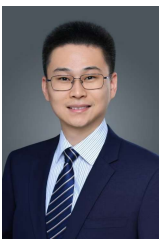
**Rongqiang zhao** received the Ph.D. degree in Harbin Institute of Technology, Harbin, China, in 2018.

He is currently an Associate Professor with the Faculty of Computing, Harbin Institute of Technology, Harbin 150001, China, and also with the National Key Laboratory of Smart Farm Technologies and Systems, Harbin Institute of Technology, Harbin 150001, China.